\documentclass[letterpaper, 11 pt, conference]{ieeeconf}  % Comment this line out
\usepackage{graphicx} %package to manage images
\usepackage{multirow}
\usepackage{booktabs}
\usepackage[font=small]{caption}
\usepackage{subcaption}

\IEEEoverridecommandlockouts                              
\usepackage[utf8]{inputenc}
\usepackage[T1]{fontenc}

\title{\LARGE \bf Why Do Students Drop Out? University Dropout Prediction and Associated Factor Analysis Using Machine Learning Techniques}

\author{ Sean Kim$^{*1, 2}$, Eliot Yoo$^{*1, 3}$, and Samuel Kim$^{1}$ 
\\ \\
$^1$ IF Research Lab, La Palma, CA, USA \\
$^2$ Oxford Academy, Cypress, CA, USA \\
$^3$ Cypress High School, Cypress, CA, USA 
\\
{\tt\small \{seankim.hahjean, eliot4yoo\}@gmail.com} \\{\tt\small sam@ifresearchlab.com}
\thanks{$^{*}$ These authors contributed equally. The names are listed in alphabetical order.}
}

\begin{document}

\maketitle
\begin{abstract}
Graduation and dropout rates have always been a serious consideration for educational institutions and students. High dropout rates negatively impact both the lives of individual students and institutions. To address this problem, this study examined university dropout prediction using academic, demographic, socioeconomic, and macroeconomic data types. Additionally, we performed associated factor analysis to analyze which type of data would be most influential on the performance of machine learning models in predicting graduation and dropout status. These features were used to train four binary classifiers to determine if students would graduate or drop out. The overall performance of the classifiers in predicting dropout status had an average ROC-AUC score of 0.935. The data type most influential to the model performance was found to be academic data, with the average ROC-AUC score dropping from 0.935 to 0.811 when excluding all academic-related features from the data set. Preliminary results indicate that a correlation does exist between data types and dropout status. 
\end{abstract}

\section{INTRODUCTION}
Student dropout represents a significant and complex issue within higher education \cite{Kim2018}.  Not only does it interfere with educational development, but dropping out of college also has consequences for individual students and institutions. At the individual level, there is a waste of personal resources, time, and money and a loss in employment prospects and earnings potential \cite{Sarra2018}. On the other hand, college graduates tend to have significantly higher personal income and happiness and lower depression and stress levels \cite{Faas2018}. At the institutional level, universities are negatively impacted educationally and economically \cite{Sarra2018}. Due to the various negative repercussions of university dropout, improving student retention would benefit individual students and educational institutions. Predicting which students are likely to drop out or graduate accurately would help alleviate dropout rates in higher education. By identifying at-risk students most likely to be at risk of dropping out, more resources can be focused on them, which may lead to improved retention and graduation rates. 

Studies have examined a complex array of unique factors that include more than just academic information to examine dropout in higher education, including socio-demographic, economic, and mental factors \cite{Sarra2018, Faas2018, Tinto1975, ONeill2011}. Incorporating a wide array of relevant factors is vital because non-academic variables have been indicated to be involved with school dropout \cite{ROSENTHAL1998}. Common predictors examining students’ background info include demographics, socioeconomic status, prior academic experience, performance, and student-related records such as age, gender, and status \cite{Kabathova2021}.

Various studies have studied the effects of specific predictors for university dropout but generally have little consensus on what specific factors are important \cite{Tinto1975, ROSENTHAL1998, Sosu2019,  Voelkle2008}. According to \cite{ONeill2011}, a wide range of different entry qualifications are associated with greater chances of student dropout. There was no specific pattern of demographic variables particularly important to drop out. On the other hand, the effects of socioeconomic, psychological, and educational variables needed to be better investigated further. 

Our study aims to examine the impact that demographic, socioeconomic, academic, and macroeconomic data have on predicting student dropout in university with different machine-learning techniques. 

\section{Procedure}
\subsection{Data Source}
We utilized a data set collected in another study related to dropout prediction, which consists of records related to students with 17 different undergraduate degrees from 2008 to 2019. The data set was created using data from the Polytechnic Institute of Portalegre as well as data from the General Directorate of Higher Education and the Contemporary Portugal Database. The information includes demographic data, socioeconomic data, macroeconomic data, data at the time of student enrollment, and academic data at the end of semesters. There are 4424 total records with 35 attributes. See \cite{Realinho2022} for further information regarding the specifics of the data set. 

\subsection{Exploratory Data Analysis}

\begin{figure}[t!]
  \centering
      \includegraphics[width=1\linewidth]{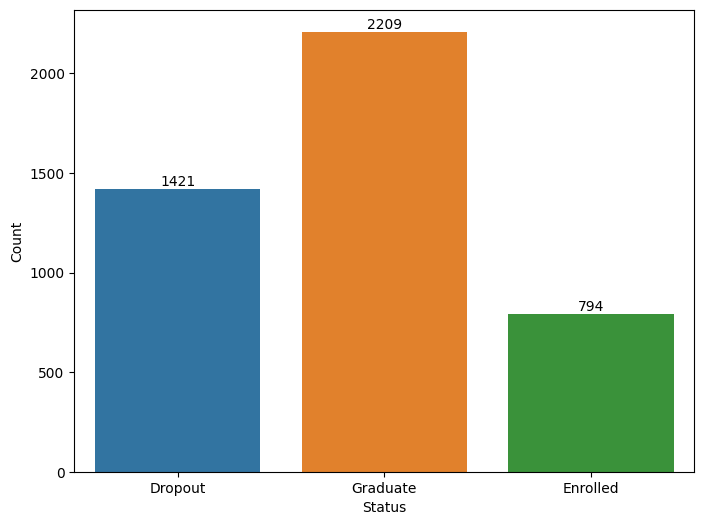}
  \caption{Class distribution of original data set.}
  \label{fig:three features}
\end{figure}

\begin{figure}[t!]
  \centering
      \includegraphics[width=1\linewidth]{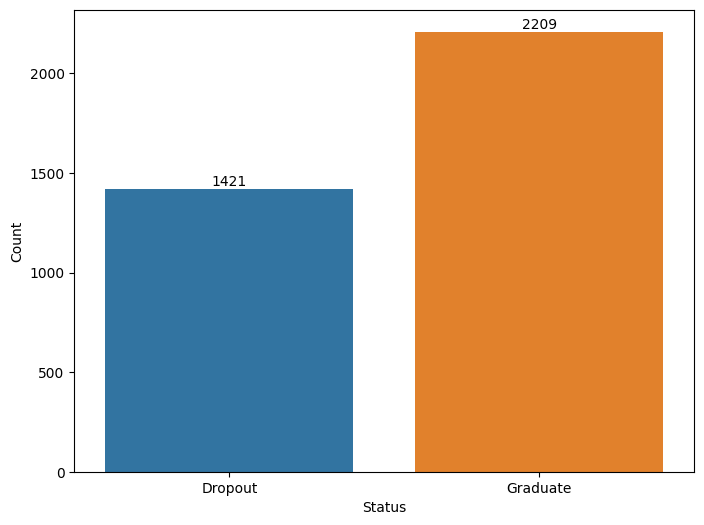}
  \caption{Class distribution after the ‘Enrolled’ class was dropped and represents the final class distributions used in the study. 
}
  \label{fig:dropout v graduate}
\end{figure}

\begin{figure}[t!]
  \centering
      \includegraphics[width=1\linewidth]{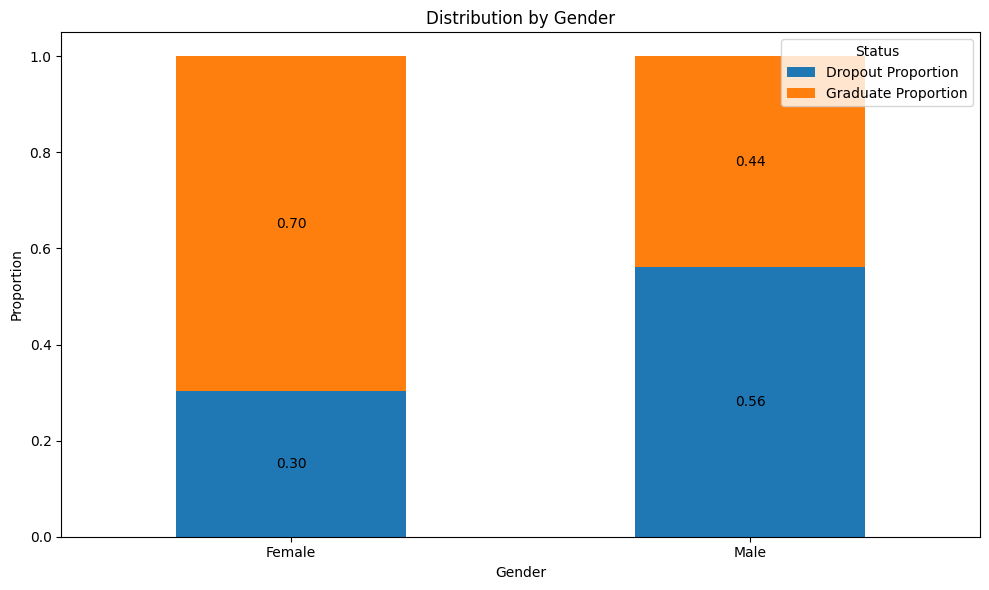}
  \caption{Distribution of dropouts and graduates by gender.}
  \label{fig:gender}
\end{figure}

\begin{figure}[t!]
  \centering
      \includegraphics[width=1\linewidth]{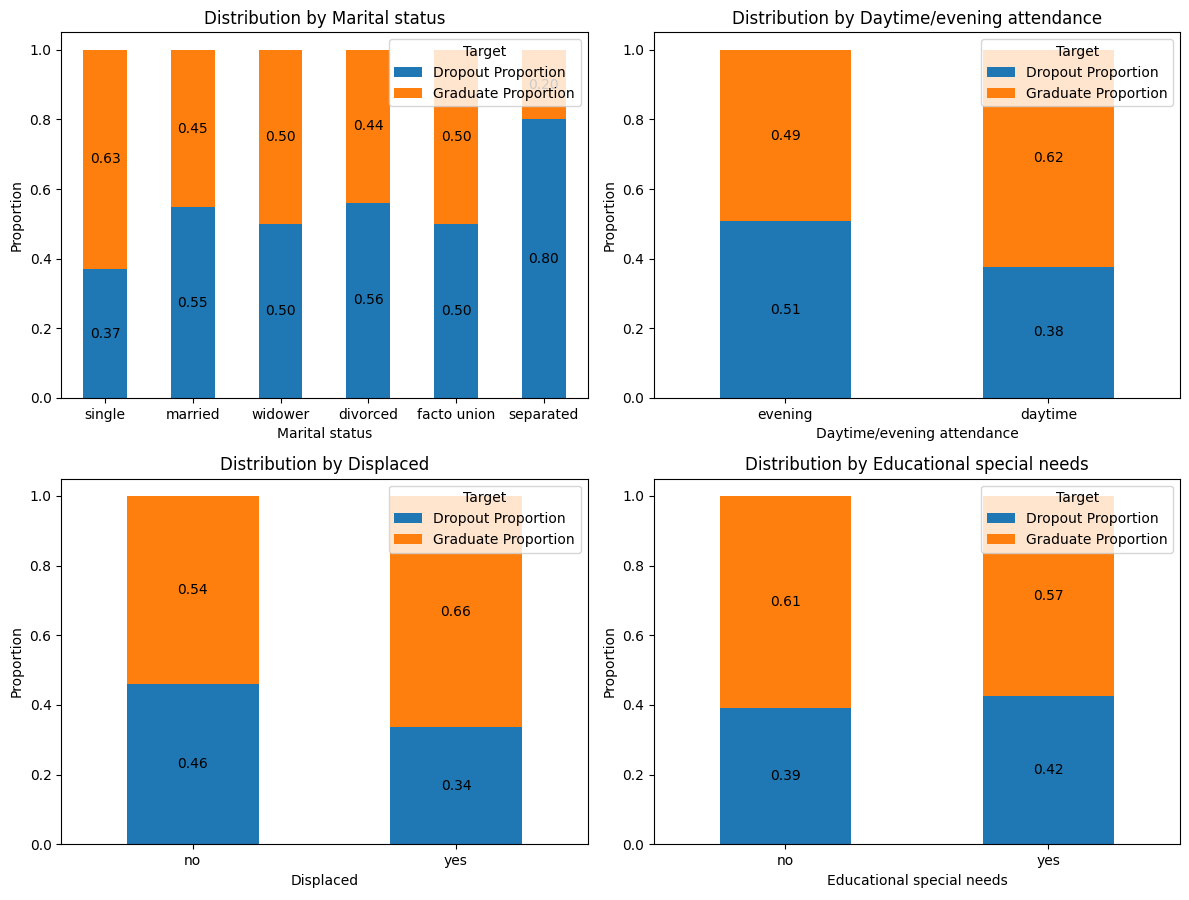}
  \caption{Distribution of dropouts and graduates by marital status, daytime/evening attendance, displaced status, and educational special needs status. }
  \label{fig:half1}
\end{figure}

\begin{figure}
    \centering
    \includegraphics[width=1\linewidth]{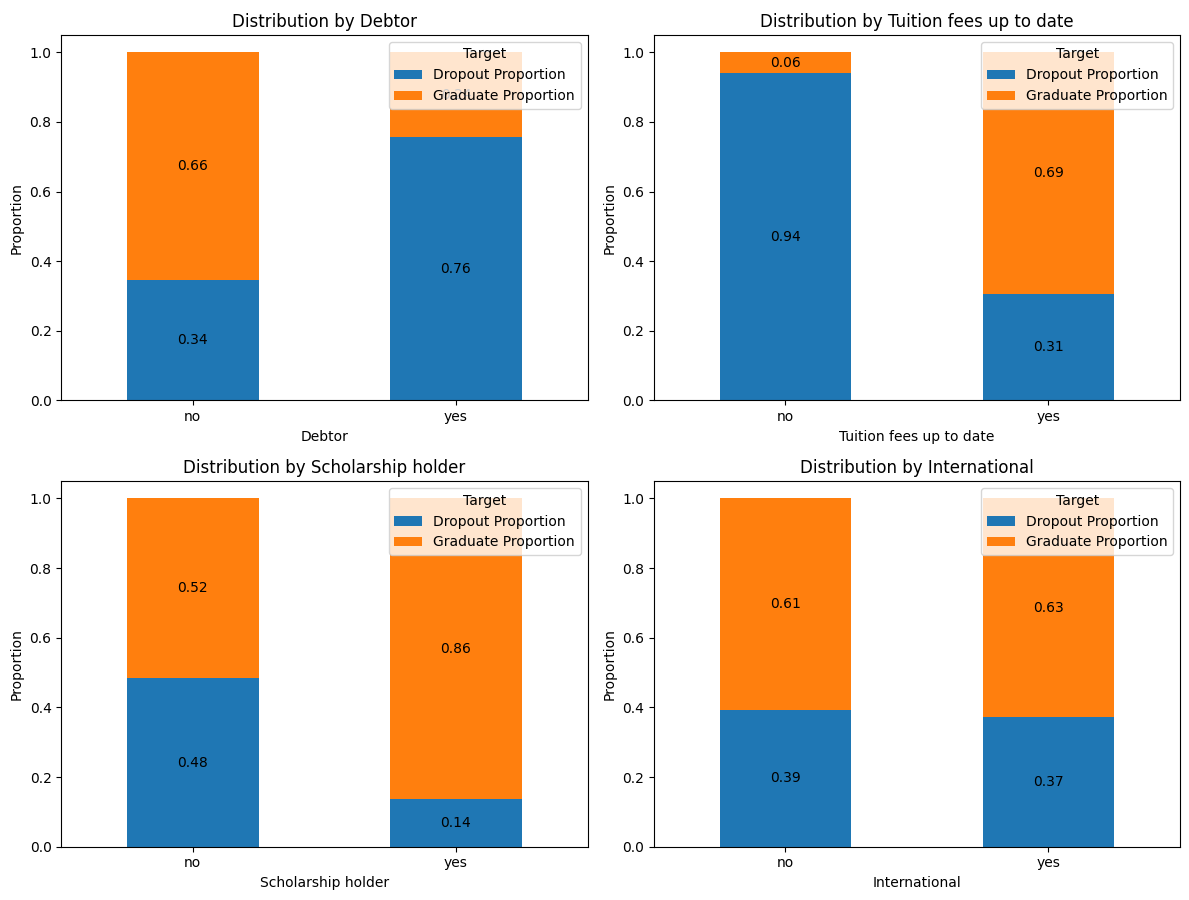}
    \caption{Distribution of dropouts and graduates by debtor status, tuition fee status, scholarship holder status, and international student status. }
    \label{fig:half2}
\end{figure}

Of the 4424 total records of students, 1421 had the status of dropout, 2209 had the status of graduate, and 794 were enrolled, as shown in Figure~\ref{fig:three features}. We implemented a binary classification task by focusing on the 2209 graduates and 1421 dropouts for 3630 total records in our study, removing the 794 enrolled students from our data due to ambiguous status. The final class distribution of our data is shown in Figure \ref{fig:dropout v graduate}. 2381 students were females, while 1249 students were male. Figure \ref{fig:gender} depicts the distribution of dropouts and graduates across the two genders, which reveals that females tended to be more academically successful than males. Figure \ref{fig:half1} and Figure \ref{fig:half2} show the distribution of dropouts and graduates across marital status, daytime or evening attendance, displacement status, educational special needs, debtor status, whether tuition fees were up to date, scholarship holders, and international student status. Features related to socioeconomic status, such as late tuition fees and debtor status, seemed to correlate highly with dropout status. Dropout rates were highest for students with late tuition fees, with approximately 94 percent of such students dropping out. 76 percent of debtors were dropouts. On the other hand, 86 percent of scholarship holders became graduates, which may be due to a higher desire to succeed and confirms existing findings about the relationship between scholarships and student success \cite{Ganem2011}. Concerning marital status, single students appeared more likely to graduate, while legally separated students were likelier to drop out. 

\begin{figure}[t!]
  \centering
      \includegraphics[width=1\linewidth]{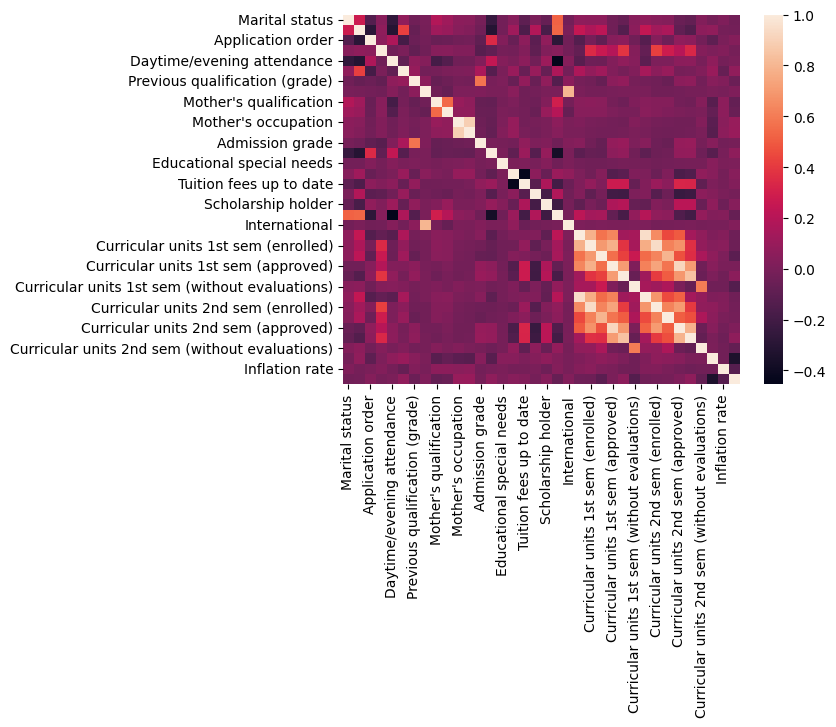}
      \caption{Correlation Heat maps for all features.} 
      \label{fig:heatmap}
\end{figure}

Correlation heat maps were created to analyze relationships between different features. Figure \ref{fig:heatmap} graphs all academic, demographic, socioeconomic, and macroeconomic features. There appears to be some correlation between certain groups such as “Nationality” and “International,” “Mother’s occupation” and “Father’s occupation,” and “Curricular units 1st sem (enrolled)” and “Curricular units 2nd sem (enrolled)” which is consistent with previous findings \cite{Realinho2022}. 

\section{Experiments}
\subsection{Methodology}
Each subject had labels stating their status in their school: graduated, dropped out, or currently enrolled. As currently enrolled students in the data set could not be used to train the machine learning models to determine whether a student would graduate or drop out, they were cut out from the primary data set, leaving 3630 subjects (removed 794 subjects). The remaining students were then split randomly into testing and training data sets (the testing set is 20\%), with their labels being graduation status.
Using this data, four different machine learning classifiers were evaluated on their performance on the binary classification task of student dropout prediction.  

A secondary test was run to find the type of data most influential on the results of student dropout prediction. The four types of data were demographic, socioeconomic, macroeconomic, and academic data. Demographic data included marital status, nationality, displaced, gender, age at enrollment, and international. Socioeconomic data included father’s qualification, mother’s qualification, father’s occupation, mother’s occupation, educational special needs, debtor, tuition fees up to date, and scholarship holder. Macroeconomic data included the unemployment rate, inflation rate, and GDP. The remaining 17 features were grouped as academic data. 

We looked at the impact on performance after excluding one data group from the data set to find the most influential data type. While training the machine learning classifiers for dropout prediction, one of the four data types was excluded (For example, only using demographic, socioeconomic, and macroeconomic for one trial, then using only demographic, socioeconomic, and academic data for the next, and so on). This was repeated a total of four times, with a different group excluded each time. Thus, the performance of machine learning classifiers was evaluated on four distinct data sets, each one missing either demographic, socioeconomic, macroeconomic, or academic data. The performance of machine learning classifiers when using all four data types was used as a baseline. 

Multiple classifiers were trained on the data set to find correlations between the multiple features and their target. Using numerous classifiers allows us to validate the results and ensure a higher accuracy in results. The machine learning classifiers that were used include the Random Forest, Decision Tree (max depth = 5), Support Vector (random\_state= 42, kernel= 'linear'), and k-nearest Neighbors (n\_neighbors = 20) with their specific parameters.

\subsection{Results}
\begin{figure}[t!]
  \centering
      \includegraphics[width=8cm]{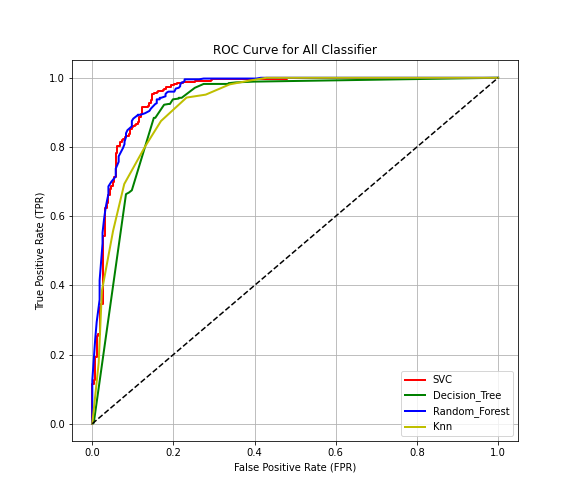}
  \caption{The ROC curve graph for the performance of all the classifiers with the whole data set.}
  \label{fig: combined ROC Curve}
\end{figure}

Receiver Operating Characteristic (ROC) Curves were graphed, and ROC-AUC (Area Under Curve) scores were used to evaluate the performance of each model. The four classifiers had ROC-AUC scores of 0.911 or higher (see Table \ref{table:auc score}). Figure \ref{fig: combined ROC Curve} shows the ROC curves for each classifier and its performance compared to each other. The best-performing classifier was the Random Forest classifier, which had a 0.955 ROC-AUC score. The classifier with the second best result was the Support Vector Classifier, with a 0.953 ROC-AUC in its performance to predict the students’ outcomes. The K-Nearest Neighbors Classifier had a ROC-AUC score of 0.92, and the Decision Tree Classifier had a 0.911 ROC-AUC score. These show a relatively good performance in predicting the students’ chances of graduating or dropping out. We can assume a positive relationship between the features of the subjects mentioned earlier and their performance in school. The numerous classifiers help confirm this as they had similar accurate performances that show these results were not all random.

\begin{figure*}[t!]
\centering
  \begin{subfigure}[]{0.4\linewidth}
    \includegraphics[width=\linewidth]{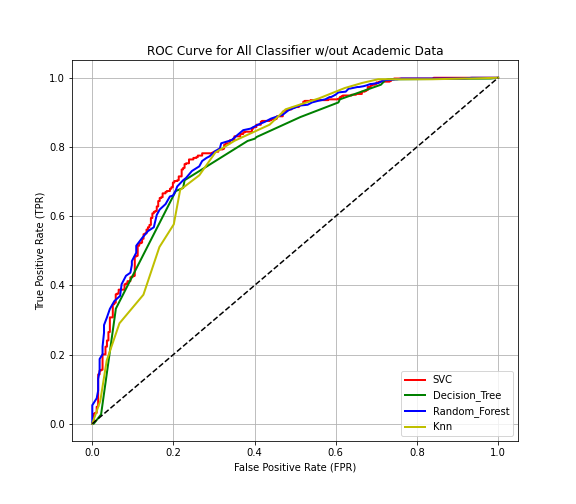}
     \caption{ROC Curve with Academic Data Excluded}     
  \end{subfigure}
  \begin{subfigure}[]{0.4\linewidth}
    \includegraphics[width=\linewidth]{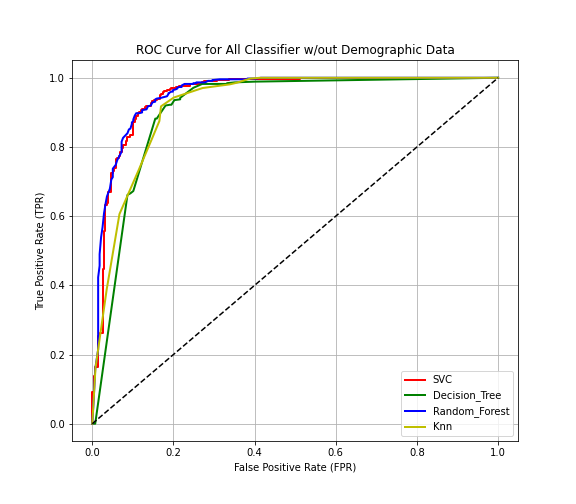}
     \caption{ROC Curve with Demographic Data Excluded}
  \end{subfigure}
  \begin{subfigure}[]{0.4\linewidth}
    \includegraphics[width=\linewidth]{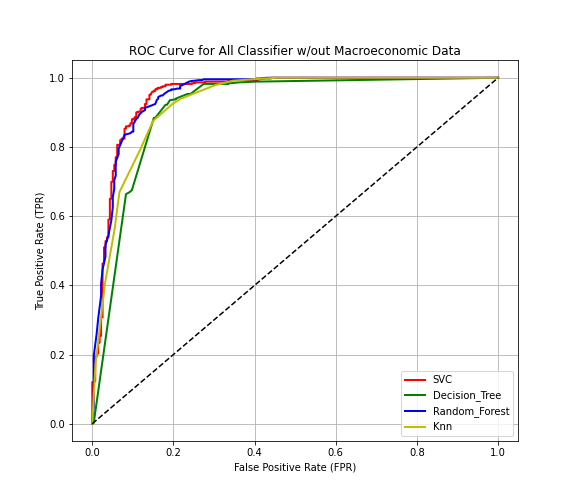}
    \caption{ROC Curve with Macroeconomic Data Excluded}
  \end{subfigure}
  \begin{subfigure}[]{0.4\linewidth}
    \includegraphics[width=\linewidth]{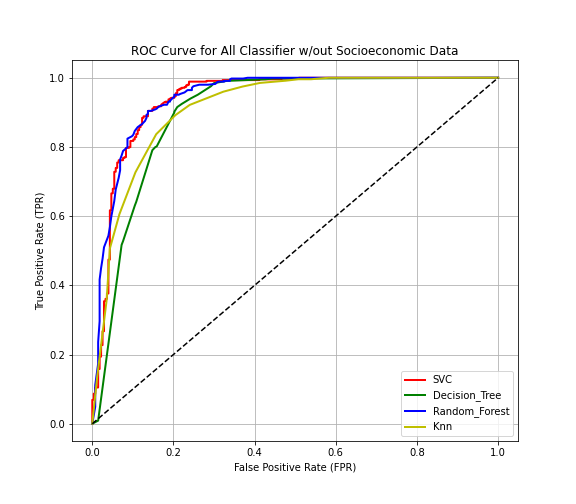}
     \caption{ROC Curve with Socioeconomic Data Excluded}     
  \end{subfigure}
  \caption{ROC curve graphs for each of the tests, with each excluding one different category.}
  \label{fig: combined roc graphs}
\end{figure*}

We found that the academic data was the most influential to the results as the average ROC-AUC score dropped 0.124 (see Table \ref{table:separated auc score} for all AUC scores for each classifier), lowering the scores significantly more than the other four tests. This is also evident in Figure \ref{fig: combined roc graphs}, as the ROC curves are considerably flatter for academic data than the other tests. These results are reasonable as the academic data type accounted for 17 of the 34 total features, which is considerably greater than the other 3 data types and is a limitation of this study. The second most influential data type was the socioeconomic data, with its classifiers’ average ROC-AUC score dropping 0.013, followed by the demographic data, with the average ROC-AUC score dropping 0.001, and the least influential features was macroeconomic data, with the average ROC-AUC score of 0.935, not dropping from the overall test. Thus, our results indicate that while academic data significantly impacted the performance of the machine learning classifiers, macroeconomic data had little to no impact. 
\begin{table}[t!]
\centering
\begin{tabular}{c|c|c|c|c|c|c}
\hline 
 & SVC & DT & RF & KNN & Average & STDV\\
\hline \hline
\multirow{1}{*}{ROC-AUC} & 0.953 & 0.911 & 0.955 & 0.92 & 0.935 & 0.023\\
\hline
\end{tabular}
 \caption{The AUC scores for all of the classifiers, including their average and standard deviation}
\label{table:auc score}
\end{table}

\begin{table*}[t!]
\centering
\begin{tabular}{c|c|c|c|c}
\hline 
 & Excl. Academic Data & Excl. Demographic Data & Excl. Macroeconomic Data & Excl. Socioeconomic Data\\
\hline \hline 
SVC & 0.821 & 0.950 & 0.952 & 0.939\\
\hline
DT & 0.798 & 0.912 & 0.904 & 0.894\\
\hline
RF & 0.823 & 0.953 & 0.954 & 0.945\\
\hline
KNN & 0.80 & 0.92 & 0.93 & 0.91\\
\hline
Average & 0.811 & 0.934 & 0.935 & 0.922\\
\hline
STDV & 0.013 & 0.208 & 0.023 & 0.024\\
\hline
\end{tabular}
 \caption{AUC scores for each of the four tests including their averages and standard deviations}
\label{table:separated auc score}
\end{table*}

\subsection{Feature Importance}
Calculating feature importance is a crucial aspect of machine learning tasks as it allows for a clear understanding of what features may be contributing most to the final decision made by models. Through feature importance, future work can gain insight into types of features that may be more suitable for predicting dropout and graduation in higher academia. Feature importance was found for the best-performing classifier from our experiments, the Random Forest Classifier. The top three features considered most important were: “Curricular units 1st sem (approved)”, “Curricular units 2nd sem (approved)”, and “Curricular units 2nd sem (grade)”. 

\subsection{Limitations}
A limitation of this study was the imbalanced number of features in each data category. For example, academic data contained 17 features, while macroeconomic data only included 3 features. The imbalanced features may have affected the study results and should be considered in future work. 

\section{Conclusion}
This study aimed to explore the connections between a student’s academic, demographic, socioeconomic, and macroeconomic data and their chances of graduating or dropping out and analyzing the impact of each data type on the results. Various machine learning classifiers were used to tackle the binary classification task of dropout prediction. Of the classifiers trained, Random Forest had the best performance, and academic data was found to have the highest impact on performance for the task of dropout prediction. This research can be applied in future studies predicting dropout for students to train more accurate machine learning models and obtain relevant and impactful data for this task. Dropout prediction can help identify at-risk students at an early stage, which can help determine where to divert additional resources to reduce dropout rates. 

\bibliographystyle{IEEEbib}
{\bibliography{mybib}}

\end{document}